\documentclass[11pt]{article}

\usepackage[preprint]{acl}

\usepackage{times}
\usepackage{latexsym}
\usepackage{amsmath}

\usepackage{natbib}

\usepackage[T1]{fontenc}

\usepackage[utf8]{inputenc}

\usepackage{microtype}

\usepackage{inconsolata}

\usepackage{graphicx}

\usepackage{caption}

\usepackage{makecell}
\usepackage{booktabs}
\usepackage{multirow}

\usepackage{algorithm}
\usepackage{algorithmic}
\usepackage{cleveref}

\title{Ensemble Self-Training for Unsupervised Machine Translation}

\author{
Ido Aharon$^1$ \quad Jonathan Shaki$^1$ \quad Sarit Kraus$^1$ \\
$^1$Bar-Ilan University
}

\begin{document}
\maketitle

\begin{abstract}
We present an ensemble-driven self-training framework for unsupervised neural machine translation (UNMT). Starting from a primary language pair, we train multiple UNMT models that share the same translation task but differ in an auxiliary language, inducing structured diversity across models. We then generate pseudo-translations for the primary pair using token-level ensemble decoding, averaging model predictions in both directions. These ensemble outputs are used as synthetic parallel data to further train each model, allowing the models to improve via shared supervision. At deployment time, we select a single model by validation performance, preserving single-model inference cost. Experiments show statistically significant improvements over single-model UNMT baselines, with mean gains of 1.7 chrF when translating from English and 0.67 chrF when translating into English.
\end{abstract}

\section{Introduction}

Neural machine translation (NMT) performs well when large parallel corpora are available, but many language pairs---especially low-resource ones---lack such supervision \citep{stahlberg2020neural,forcada2017making,koehn2017six,ranathunga2023neural}. Unsupervised neural machine translation (UNMT) addresses this setting by learning from monolingual data alone \citep{artetxe2019effective,ranathunga2023neural}. Despite substantial progress through back-translation, denoising objectives, and multilingual pretraining \citep{liu2020multilingual,chauhan2022fully,ataman2025machine}, UNMT remains fragile: training quality is highly sensitive to initialization, and self-training can reinforce poor pseudo-translations rather than correct them \citep{marie2018unsupervised,kim2020and,lin2025addressing,arazo2020pseudo,he2019revisiting}.

The central mechanism in UNMT is back-translation \citep{forcada2017making,ranathunga2023neural}, which turns monolingual sentences into synthetic parallel data by translating in one direction and training the reverse direction to reconstruct the original sentence. This enables learning without gold bitext, but it also creates a closed feedback loop: each model is trained on pseudo-labels that it generated itself, or that were generated by a closely related counterpart. When those pseudo-translations are noisy or degenerate, errors and biases can compound over training \citep{shumailov2024ai,shaki2023cognitive,trabelsi2026pro}. In difficult settings, such as weak cross-lingual alignment or limited target-language coverage, individual models may converge to poor equilibria rather than to robust translation behavior \citep{marie2018unsupervised,kim2020and,lin2025addressing}.

In this work, we address this limitation by replacing single-model self-training with \emph{ensemble-driven self-training}. Our key idea is to use multiple UNMT models to generate pseudo-parallel data jointly, rather than relying on any one model's outputs alone. To obtain meaningful diversity, we train several models on the same primary language pair $(A,B)$, but assign each model a distinct auxiliary language $C_i$. Each model, therefore, learns the same core translation task while being exposed to different cross-lingual signals. We then ensemble these models at the token level to generate pseudo-translations for the primary pair, and use those ensemble outputs as supervision for further training. In this way, the ensemble is not merely an inference-time trick: it becomes a mechanism for improving the training signal itself.

Our framework separates \emph{training-time collaboration} from \emph{deployment-time efficiency}. Multiple models collaborate only while generating synthetic supervision; after training, we deploy a single model selected by validation performance. This preserves the inference cost of standard UNMT while allowing training to benefit from more stable and higher-quality pseudo-labels.

We evaluate the method in fully unsupervised translation settings spanning both low-resource and high-resource scenarios. In addition, when training decoder-only LLMs for UNMT from weak initialization, we identify two practical failure modes: (1) the model copies the input instead of translating it, and (2) the model produces nearly the same output distribution regardless of the input. To mitigate these issues, we introduce a mixed-training strategy that stabilizes early learning. Across experiments, ensemble-driven self-training yields consistent gains over single-model baselines, with the largest improvements occurring in the more difficult translation direction. These results suggest that structured model diversity and shared pseudo-supervision provide a simple and effective way to improve UNMT without increasing deployment-time complexity.

Our contributions are as follows:
\begin{itemize}
    \item We analyze collapse modes that arise when training next-token-prediction LLMs for unsupervised translation from weak initialization, and introduce a mixed-training strategy that stabilizes learning (\Cref{sec:low-resource-mono}).
    \item We propose an ensemble-driven self-training framework for UNMT in which multiple models learn from ensemble-generated pseudo-parallel data while differing only in their auxiliary language (\Cref{sec:methodology}).
    \item We show that ensemble-generated training yields consistent improvements over single-model UNMT baselines while retaining single-model inference efficiency (\Cref{sec:results}).
\end{itemize}

\subsection{Related Work}

\paragraph{UNMT with Decoder-Only LLMs}
Early UNMT methods relied on architectures and objectives designed specifically for unsupervised translation \citep{artetxe2017unsupervised,artetxe2019effective}. More recent work has explored using standard generative LLMs for translation with little or no parallel supervision \citep{han2021unsupervised,garcia2023unreasonable,ataman2025machine}. This shift makes it possible to build UNMT systems directly on top of broadly available pretrained models, which is the setting we consider here.

\paragraph{Ensemble Learning}
Ensembling has recently been studied extensively for LLMs \citep{chen2025harnessing,mienye2025ensemble,ashiga2025ensemble}. Common approaches include token-level aggregation of predictive distributions \citep{jin2024collaborative,yu2024breaking}, routing inputs to specialized models \citep{zhao2024eagle,cai2025survey}, averaging weights across models \citep{shaki2025out} and majority-style answer selection after reasoning \citep{li2024more,chen2025harnessing}. In unsupervised translation, however, routing is nontrivial and explicit reasoning is typically absent, making token-level ensembling the most natural choice.

\paragraph{Ensemble Methods for Machine Translation}
Ensemble methods have shown gains in supervised NMT \citep{garmash2016ensemble,wang2020transductive,man2024ensemble}, but, to our knowledge, they have not been used as a \emph{training-time supervision mechanism} in UNMT. This distinction is important: in supervised settings, ensembles are typically used at inference time, whereas in our setting, ensemble outputs can be recycled into additional synthetic supervision. Our method leverages this property while still avoiding multi-model deployment.

\paragraph{Evaluation}
We evaluate translation quality primarily using chrF \citep{popovic2015chrf}, a character n-gram F-score that has been shown to correlate well with human judgments \citep{smith2016climbing,freitag2020bleu,mathur2020tangled}, particularly for morphologically rich languages. In addition, we report BLEU \citep{papineni2002bleu,reiter2018structured,saadany2021bleu} scores using sacreBLEU \citep{post2018call}\footnote{Signature: \texttt{nrefs:1|case:mixed|eff:no|tok:13a|\\smooth:exp|version:2.5.1}.} to facilitate comparison with prior UNMT work \citep{han2021unsupervised,chronopoulou2020reusing,artetxe2017unsupervised,artetxe2019effective,lin2025addressing}. Finally, we report COMET \citep{rei2020comet,guerreiro2024xcomet}\footnote{Using XComet-XXL as evaluator.} as a neural evaluation metric that has been shown to better capture semantic adequacy. chrF serves as our primary evaluation metric, while BLEU and COMET are provided as complementary indicators of translation quality.

\section{Method}
\label{sec:methodology}

\subsection{Overview}

Our approach proceeds in three stages. First, we independently train several UNMT models that all learn translation for the main language pair $(A,B)$, but each additionally incorporates a distinct auxiliary language $C_i$. The presence of different auxiliary languages encourages diversity across models.

Second, we combine the predictions of these models during decoding using token-level ensembling. The resulting ensemble translations are treated as pseudo-parallel data for the main language pair. Each model is then further trained on this synthetic data, enabling iterative improvement through collective supervision.

Finally, we select a single model for deployment based on validation performance. In this way, the ensemble is used only during training to improve the quality of supervision, while inference remains as efficient as standard single-model translation.

The overall algorithm is summarized as \Cref{alg:ensemble_unmt}.

\begin{algorithm}[t]
\caption{Ensemble-Driven UNMT Training}
\label{alg:ensemble_unmt}
\begin{algorithmic}[1]

\STATE Initialize $N$ translation models $\{M_1, \dots, M_N\}$
\STATE Assign a distinct auxiliary language $C_i$ to each model

\FOR{each model $M_i$}
    \STATE Train $M_i$ on languages $(A,B,C_i)$ using unsupervised objectives
\ENDFOR

\FOR{$k$ steps}
    \STATE Generate pseudo-translations for $(A,B)$ using ensemble decoding
    \STATE Construct synthetic parallel dataset $\mathcal{D}_{pseudo}$
    
    \FOR{each model $M_i$}
        \STATE Train $M_i$ on $\mathcal{D}_{pseudo}$
    \ENDFOR
\ENDFOR

\STATE Select best model $M^*$ based on validation performance
\STATE Deploy $M^*$ for inference

\end{algorithmic}
\end{algorithm}

\subsection{Problem Setup}

We consider a translation task between two languages $A$ and $B$ for which no parallel data are available. Instead, we assume access only to monolingual corpora $D_A$ and $D_B$.

To introduce diversity among models, we train multiple systems that each include an additional auxiliary language. Specifically, for each model $i$, we select a distinct language $C_i$. Model $i$ is trained on translation between the three languages $\{A,B,C_i\}$ using unsupervised objectives.

Formally, model $i$ is trained on the set of translation directions
\[
(A \leftrightarrow B), \quad (B \leftrightarrow C_i), \quad (C_i \leftrightarrow A).
\]

All models, therefore, learn the same primary translation task $(A,B)$ while incorporating different auxiliary linguistic signals. This structured diversity is important for effective ensemble learning.

\subsection{Base Model Training}

Before applying the ensemble procedure, each model is trained independently using monolingual data. Training proceeds in three phases designed to stabilize learning and prevent degenerate solutions that can arise in unsupervised translation.

\subsubsection{Monolingual Training}
\label{sec:mono-training}

In the first phase, the model is trained using standard causal language modeling on monolingual sentences. Each sentence is prefixed with a language identifier indicating the target language. The objective is next-token prediction over the sentence tokens, while language identifiers are masked from the loss.

More concretely, the model is given the input \texttt{"<Tag><Sentence>"}, where the tag is \texttt{"[English/French/etc] version:"} and the sentence is the target-language sentence.
For example, a training instance might be \texttt{"English version: The quick brown fox jumps over the lazy dog."}
The loss is computed only over the sentence tokens (i.e., \texttt{"The quick brown fox jumps over the lazy dog"}), not over the language tag.

This stage allows the model to learn the basic token distribution and generation patterns of each language before attempting translation.

\subsubsection{Mixed Training}
\label{sec:mix-training}

In the second phase, training alternates between back-translation and randomized pairing tasks.

Back-translation converts monolingual sentences into synthetic parallel pairs. Given a sentence $y \sim D_B$, the model translates $y$ to $\hat{x}$ in language $A$, which is then treated as a pseudo-source sentence, used to train the model to reconstruct $y$ given $x$ only, when translating from $A \to B$.
We use the same language tags from \Cref{sec:mono-training}.

More concretely, we prompt the model with \texttt{"<source language tag><source sentence><target language tag>"} and decode a target-language translation, \texttt{<target language translation>}.
We then train the model to reconstruct the original source sentence conditioned on this translation by feeding \texttt{"<target language tag><target language translation><source language tag><source sentence>"}.
The loss is computed only over \texttt{<source sentence>}.

However, pure back-translation can lead to collapse modes such as copying the input sentence or producing constant outputs. To mitigate this, we occasionally introduce randomized sentence pairs from the source and target languages. These randomized pairs encourage the model to generate tokens in the specified target language.

In practice, we use a mixture dominated by back-translation, with a small proportion of randomized pairs. This combination stabilizes training while preserving the translation objective.

\subsubsection{Pure Back-Translation}

After the model has learned stable language representations, we transition to pure back-translation. In this stage, translation quality improves through iterative generation and reconstruction of pseudo-parallel data. Sentences from each language are translated into the other language and then used to train the reverse direction, gradually improving the translation consistency and quality between the two languages.

\subsection{Ensemble-Driven Self-Training}

Once all models have been trained independently, we introduce an ensemble stage to further improve translation quality.

\subsubsection{Ensemble Decoding}

During decoding, the predictions of multiple models are combined using token-level ensembling. At each decoding step, the models produce logits over the vocabulary. We aggregate these predictions by averaging the logits before applying the softmax function.

Formally, let $z_i$ denote the logits predicted by model $i$. The ensemble logits are computed as

\[
z_{\text{ens}} = \frac{1}{N} \sum_{i=1}^{N} z_i,
\]

and the final token distribution is obtained by applying softmax to $z_{\text{ens}}$.

Logit averaging allows models that assign high confidence to a token to influence the ensemble prediction more strongly, while still benefiting from the collective knowledge of all models.

\subsubsection{Pseudo-Parallel Data Generation}

Using ensemble decoding, we generate pseudo-translations for sentences in the monolingual corpora. These translations are paired with their original sentences to form synthetic parallel data for the main language pair $(A,B)$.

All models are then trained on this shared pseudo-parallel dataset. Importantly, the ensemble is used only to generate targets; the model parameters remain separate and are updated independently.

\subsubsection{Iterative Self-Training}

The ensemble training stage proceeds iteratively. After generating pseudo-translations using the ensemble, each model is updated using the resulting synthetic data. The improved models are then used to generate new ensemble translations in the next iteration.
This iterative process allows the models to collectively refine the quality of the pseudo-parallel data, leading to progressively better translation performance.

\subsection{Deployment}
Although multiple models participate during training, the final system uses only a single model at inference time. After the ensemble training stage completes, we evaluate all models and select the best-performing model for deployment.

\section{Experimental Setup}
\label{sec:experiments}

We evaluate the effectiveness of our method in two settings. In the first, we simulate translation between a high-resource and a low-resource language. In the second, we consider two high-resource languages with abundant monolingual data but no parallel supervision. We show that the proposed framework improves performance in both settings. For both settings, we train 10 individual models, and then test ensemble sizes of 3, 6, and 10.

Because each model requires a distinct auxiliary language, we used the Europarl dataset\footnote{https://huggingface.co/datasets/Helsinki-NLP/europarl}, which provides a large number of languages. Although Europarl is distributed as a parallel corpus, we use it only as a source of monolingual text for each language and ignore the sentence alignments. Sentences are sampled independently for each language, and aligned sentence pairs are never used jointly during training. The test and validation sets are disjoint from the training data.

\paragraph{Low-Resource Translation} We simulate translation between a high-resource and a low-resource language by selecting one language that the pretrained model handles well and one that it handles poorly, then training translation between them. Specifically, we use \texttt{Llama-3.2-1B}, with English as the high-resource language and Polish as the low-resource language, using only 100K Polish sentences for training. For auxiliary languages, we select languages that were not explicitly included during model pretraining (as stated in the model card\footnote{https://huggingface.co/meta-llama/Llama-3.2-1B}), to maximize diversity across models.

\paragraph{High-Resource Translation} We study translation between two high-resource languages using a pretrained LLM trained on both languages. In particular, we use \texttt{Qwen3-0.6B-Base} and \texttt{Qwen3-1.7B-Base}, and focus on translation between English and French.

\paragraph{Scope of the Baseline Framework.}
Our goal is to evaluate the proposed ensemble-driven self-training strategy in a minimal unsupervised translation framework in order to isolate its effect. We therefore use a relatively simple UNMT setup rather than integrating multiple advanced techniques simultaneously. Importantly, the proposed method operates purely at the level of pseudo-parallel data generation and does not depend on the specific model architecture or training objective. As a result, it is orthogonal to many existing improvements in UNMT, such as denoising objectives \citep{chauhan2022fully,liu2020multilingual} or quality-aware filtering of synthetic data \citep{xu2019improving,khatri2020filtering}. In principle, the approach could be incorporated into stronger systems by replacing their pseudo-parallel data generation step with ensemble decoding. We leave large-scale integration with state-of-the-art UNMT systems for future work.

\paragraph{Control Baseline} To enable a fair comparison, we include a matched single-model continuation baseline. Starting from the best pre-ensemble model, we perform the same amount of additional training as in the ensemble-training setting, with an identical training setup. The only difference is the source of pseudo-parallel data: in the single-model baseline it is generated by the single model, whereas in ensemble training it is generated by ensemble decoding over multiple models.

\section{Single Model Training}
\label{sec:single-model}

\subsection{Low-Resource Training}

\label{sec:low-resource-mono}

While prior work has trained generative models with unsupervised translation \citep{han2021unsupervised,lachaux2020unsupervised,wu2022autoformalization}, those studies typically use models that already have basic translation ability from pretraining. When extending this setup to models with weak initial translation ability for the target languages, in our experiments, pure back-translation often fails in one of two ways.

Formally, let $x \sim \mathcal{D}_X$ and $y \sim \mathcal{D}_Y$. In pure back-translation we generate
\[
\hat{y} = T_{X \to Y}(x), \qquad \hat{x} = T_{Y \to X}(y)
\]

and optimize
\begin{align*}
& \mathcal{L}^{\mathrm{BT}}_{Y \to X} = -\log p_{Y \to X}(x \mid \hat{y}) 
& \\
& \mathcal{L}^{\mathrm{BT}}_{X \to Y} = -\log p_{X \to Y}(y \mid \hat{x})
\end{align*}

The failures we find are:

\paragraph{Copying collapse.} A degenerate fixed point is $T_{X \to Y}(x)=x$ and $T_{Y \to X}(y)=y$, yielding
\begin{align*}
& \mathcal{L}^{\mathrm{BT}}_{Y \to X} = -\log p_{Y \to X}(x \mid x)
& \\
& \mathcal{L}^{\mathrm{BT}}_{X \to Y} = -\log p_{X \to Y}(y \mid y)
\end{align*}
whose optimal solution remains $T_{X \to Y}(x)=x$ and $T_{Y \to X}(y)=y$, completing the fixed point.

Intuitively, back-translation teaches the model to be consistent: if $x$ is the translation of $y$, then $y$ should be the translation of $x$. A degenerate solution is the identity mapping $x=y$.

\paragraph{Constant-output collapse.}
Another failure mode is deterministic output distribution independent of the input, e.g., $\hat{y}=c_Y$ for all $x$ (and symmetrically $\hat{x}=c_X$ for all $y$). Then pseudo-inputs carry no source information:
\[
\hat{y} \perp x \quad \Rightarrow \quad I(X;\hat{Y})=0
\]
The reverse model is trained with
\[
\mathcal{L}^{\mathrm{BT}}_{Y \to X} = -\log p_{Y \to X}(x \mid c_Y)
\]
so the optimum ignores input content and tends toward unconditional language modeling on $X$:
\[
p_{Y \to X}(x \mid c_Y)\approx p_X(x)
\]
Symmetrically, $p_{X \to Y}(y \mid c_X)\approx p_Y(y)$, so training drifts away from translation.

Intuitively, if the model outputs a translation $y$ independently of the source $x$, when trained to translate $y$ to $x$, $y$ carries no information about $x$; hence, the model ignores the source sentence, and cannot recover the ability to condition on it.

\paragraph{Solution} We find that the mixed-training phase (\Cref{sec:mix-training}), where the training is done mainly on back-translation but also with random pairs of the source and target language, is an effective remedy. Intuitively, randomized pairing indicates the required output language, serving as a solution to (1), while back-translation enforces correlation between source and generated text, aimed at fixing (2). However, the ratio is critical: if back-translation dominates too strongly, a collapse to mode (1) is common; if randomized pairing is too frequent, a collapse to mode (2) becomes more likely. Empirically, a back-translation ratio in the range $90\%-98\%$ worked well, and we used $98\%$ to limit potential degradation from randomized pairs.

\subsection{High-Resource Training}
In contrast to the low-resource translation, when using models that already possess strong translation capability between the target languages, we find that starting directly with back-translation works well, in line with previous works \citep{han2021unsupervised,ataman2025machine}.

\subsection{Training Parameters}
\label{sec:single-model-params}
We use an initial learning rate of $4\cdot 10^{-6}$, and reduce it by a factor of $2$ after the first and second 2K training steps. The batch size is set to 16 to fit within GPU memory constraints. We aim to train the models until performance no longer improves; we find that 20K batches for the low-resource case and 10K for the high-resource work well. In the low-resource setting, we additionally use 1K monolingual pretraining batches followed by 2K mixed-training batches to stabilize learning. The models are trained and evaluated with a temperature of $0.1$ (a small but positive temperature was found to elicit UNMT performance in \citep{han2021unsupervised}). Parameters were manually tuned on a different dataset (WMT-16 \citep{bojar-EtAl:2016:WMT1}) using a single model (Qwen-3-0.6B-Base) for English–French translation\footnote{Hardware and budget estimates appear in \Cref{sec:training-details}. Code repository, including full logs and graphs, will be released upon publication.}.

\subsection{Results}

\begin{table}[t]
  \centering
  \small

  \begin{tabular}{clc}
    \toprule
    Model & Direction & \makecell{Average \\ Score} \\
    \midrule
    
    \multirow{2}{*}[-0.6ex]{\makecell{Llama-3.2-1B \\ (Low-resource)}}
    & EN$\to$PL & 16.60 \\
    & PL$\to$EN & 20.03 \\
    
    \midrule
    \multirow{2}{*}[-0.6ex]{\makecell{Qwen3-0.6B-Base \\ (High-resource)}}
    & EN$\to$FR & 51.29 \\
    & FR$\to$EN & 53.20 \\
    
    \midrule
    \multirow{2}{*}[-0.6ex]{\makecell{Qwen3-1.7B-Base \\ (High-resource)}}
    & EN$\to$FR & 55.79 \\
    & FR$\to$EN & 56.95 \\
    
    \midrule
    
  \end{tabular}
  \caption{Average chrF score of individual models for the main language pair, on 3000 sentences.}
  \label{tab:single-model-summary}
\end{table}

The single-model performance is provided in \Cref{tab:single-model-summary}. As expected, high-resource training performs much better than low-resource training. Even in the high-resource case, translating into English yields higher-quality translations, in line with prior work \citep{chronopoulou2020reusing,han2021unsupervised}. 
Full results per model and auxiliary language are provided in \Cref{tab:llama1-all-single,tab:qwen0.6-all-single,tab:qwen1.7-all-single}\footnote{Omitted tables can be found in the appendix.}. 

\begin{figure}[t]
  \centering
  \includegraphics[width=0.48\textwidth]{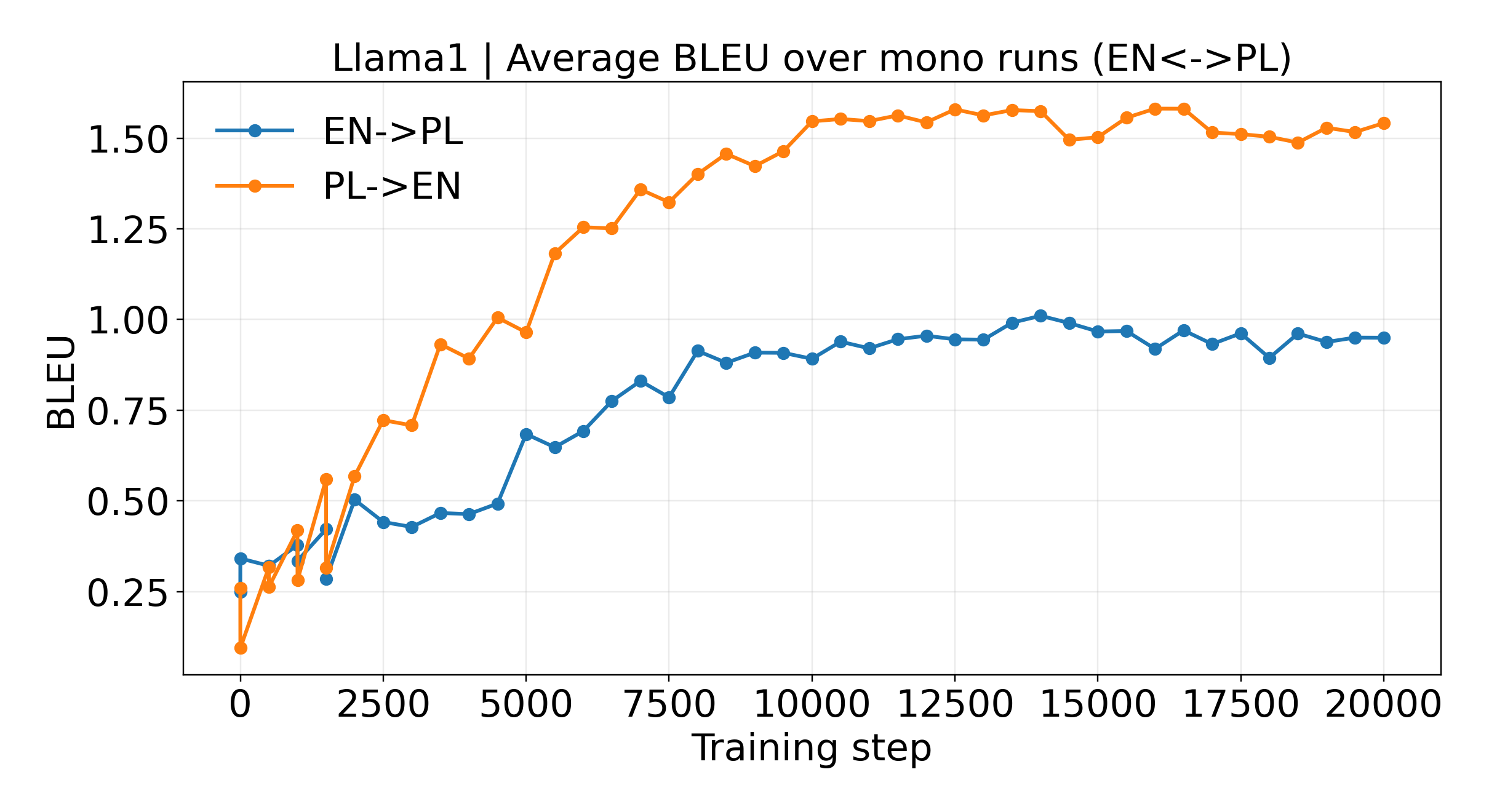}
  \caption{100 sentences BLEU for \texttt{Llama-3.2-1B}: mean performance across all single models.}
  \label{fig:llama1-single-avg}
\end{figure}

\paragraph{Training Dynamics} The average BLEU scores of the models as a function of the training steps is provided in \Cref{fig:llama1-single-avg,fig:qwen0.6-single-avg,fig:qwen1.7-single-avg}. In the high-resource case, even before any training, the models already possess some translation capabilities (around 12 BLEU for \texttt{Qwen3-0.6B-Base} and 14 for \texttt{Qwen3-1.7B-Base}). The sharp increase in the beginning likely reflects adaptation to our format and dataset. In the low-resource training, the BLEU score begins around 0.22, and increases more gradually, indicating the model indeed possesses no initial translation ability between the languages.

\section{Ensemble Training}
In this stage, we take the models trained in \Cref{sec:single-model} and continue training them using pseudo-parallel data generated by their ensemble.

\subsection{Training Parameters}

We use fixed learning rate of $10^{-6}$ (which is the learning rate used in the single model training, after the first half of the training). The same batch size, temperature, training corpus, and number of batches from the single model settings are used (\Cref{sec:single-model-params}). For an ensemble of size $k$, we select the first $k$ models according to the order shown in \Cref{tab:llama1-all-single,tab:qwen0.6-all-single,tab:qwen1.7-all-single}\footnote{Hardware and budget estimates appear in \Cref{sec:training-details}}.

\subsection{Results}
\label{sec:results}

\begin{table}[t]
\centering
\small
\setlength{\tabcolsep}{4pt}
\begin{tabular}{lccccc}
\toprule
Model & \makecell{Ensemble \\ Size} & Direction & \makecell{Best \\ Model \\ Before \\ +Single\\Training} & \makecell{Best \\ Model \\ After \\Ensemble \\ Training} \\
\midrule

\multirow[c]{6}{*}[-1.2ex]{\makecell{Llama-1B \\ (low-\\resource)}}
& \multirow{2}{*}{3} & EN$\to$PL & 18.93 & 21.44 \\
&  & PL$\to$EN & 22.80 & 23.47 \\
\cmidrule(lr){2-5}
& \multirow{2}{*}{6} & EN$\to$PL & 18.93 & 21.01 \\
&  & PL$\to$EN & 22.80 & 23.79 \\
\cmidrule(lr){2-5}
& \multirow{2}{*}{10} & EN$\to$PL & 18.93 & 21.99 \\
&  & PL$\to$EN & 22.80 & 24.25 \\

\midrule

\multirow[c]{6}{*}[-1.2ex]{\makecell{Qwen-0.6B \\ (high-\\resource)}}
& \multirow{2}{*}{3} & EN$\to$FR & 52.43 & 54.59 \\
&  & FR$\to$EN & 53.87 & 54.84 \\
\cmidrule(lr){2-5}
& \multirow{2}{*}{6} & EN$\to$FR & 52.43 & 53.97 \\
&  & FR$\to$EN & 53.87 & 54.66 \\
\cmidrule(lr){2-5}
& \multirow{2}{*}{10} & EN$\to$FR & 53.29 & 54.57 \\
&  & FR$\to$EN & 54.23 & 54.82 \\

\midrule

\multirow[c]{6}{*}[-1.2ex]{\makecell{Qwen-1.7B \\ (high-\\resource)}}
& \multirow{2}{*}{3} & EN$\to$FR & 56.67 & 57.60 \\
&  & FR$\to$EN & 57.18 & 57.50 \\
\cmidrule(lr){2-5}
& \multirow{2}{*}{6} & EN$\to$FR & 56.67 & 57.52 \\
&  & FR$\to$EN & 57.18 & 57.39 \\
\cmidrule(lr){2-5}
& \multirow{2}{*}{10} & EN$\to$FR & 56.67 & 57.57 \\
&  & FR$\to$EN & 57.28 & 57.35 \\

\midrule

\multirow[c]{3}{*}[0ex]{\textbf{Mean}}
& \multirow{3}{*}{\textbf{*}} 
& \textbf{EN$\boldsymbol{\to}$*} & \textbf{42.77} & \textbf{44.47} \\
& & \textbf{*$\boldsymbol{\to}$EN} & \textbf{44.67} & \textbf{45.34} \\
& & \textbf{Mean} & \textbf{43.72} & \textbf{44.91} \\

\bottomrule
\end{tabular}
\caption{Best individual model chrF score.}
\label{tab:ensemble-summary-chrf}
\end{table}

\begin{table*}[t]
  \centering
  \small
  \setlength{\tabcolsep}{4pt}
  \begin{tabular}{cclcccccc}
    \toprule
    \makecell{Model Name} & \makecell{Ensemble \\ Size} & Direction & \makecell{Best Model \\ Before} & \makecell{Best Model \\ After} & \makecell{Avg. Models \\ Before} & \makecell{Avg. Models \\ After} & \makecell{Ensemble \\ Before} & \makecell{Ensemble \\ After} \\
    \midrule
    \multirow{6}{*}[-0.6ex]{\texttt{Llama-3.2-1B}} & \multirow{2}{*}[-0.3ex]{3} & EN$\to$PL & 17.57 (cs) & 21.44 (cs) & 17.30 & 21.27 & 18.51 & 21.09 \\
     &  & PL$\to$EN & 22.29 (bg) & 23.47 (el) & 20.62 & 23.36 & 20.41 & 23.54 \\
    \cmidrule(lr){2-9}
     & \multirow{2}{*}[-0.3ex]{6} & EN$\to$PL & 17.56 (cs) & 21.01 (cs) & 16.97 & 20.86 & 16.77 & 20.45 \\
     &  & PL$\to$EN & 22.17 (bg) & 23.79 (cs) & 19.79 & 23.56 & 20.35 & 23.55 \\
    \cmidrule(lr){2-9}
     & \multirow{2}{*}[-0.3ex]{10} & EN$\to$PL & 17.70 (cs) & 21.99 (bg) & 16.60 & 21.45 & 17.18 & 20.97 \\
     &  & PL$\to$EN & 22.35 (bg) & 24.25 (sk) & 20.03 & 23.85 & 20.53 & 23.91 \\
    \midrule
    \addlinespace[0.3ex]
    \multirow{6}{*}[-0.6ex]{\texttt{Qwen3-0.6B}} & \multirow{2}{*}[-0.3ex]{3} & EN$\to$FR & 51.60 (bg) & 54.59 (da) & 51.29 & 54.52 & 51.57 & 54.48 \\
     &  & FR$\to$EN & 53.55 (bg) & 54.84 (bg) & 53.21 & 54.80 & 53.58 & 54.77 \\
    \cmidrule(lr){2-9}
     & \multirow{2}{*}[-0.3ex]{6} & EN$\to$FR & 51.59 (bg) & 53.97 (cs) & 51.25 & 53.94 & 51.50 & 53.94 \\
     &  & FR$\to$EN & 53.59 (bg) & 54.66 (cs) & 53.24 & 54.58 & 53.63 & 54.70 \\
    \cmidrule(lr){2-9}
     & \multirow{2}{*}[-0.3ex]{10} & EN$\to$FR & 51.98 (ro) & 54.57 (pt) & 51.29 & 54.42 & 51.61 & 54.51 \\
     &  & FR$\to$EN & 53.64 (ro) & 54.82 (pt) & 53.20 & 54.73 & 53.67 & 54.90 \\
    \midrule
    \addlinespace[0.3ex]
    \multirow{6}{*}[-0.6ex]{\texttt{Qwen3-1.7B}} & \multirow{2}{*}[-0.3ex]{3} & EN$\to$FR & 55.99 (bg) & 57.60 (bg) & 55.76 & 57.56 & 55.93 & 57.60 \\
     &  & FR$\to$EN & 57.05 (bg) & 57.50 (da) & 56.89 & 57.43 & 57.03 & 57.48 \\
    \cmidrule(lr){2-9}
     & \multirow{2}{*}[-0.3ex]{6} & EN$\to$FR & 55.93 (bg) & 57.52 (bg) & 55.75 & 57.47 & 55.98 & 57.41 \\
     &  & FR$\to$EN & 57.16 (de) & 57.39 (bg) & 56.93 & 57.36 & 57.18 & 57.42 \\
    \cmidrule(lr){2-9}
     & \multirow{2}{*}[-0.3ex]{10} & EN$\to$FR & 56.14 (bg) & 57.57 (es) & 55.79 & 57.49 & 56.06 & 57.55 \\
     &  & FR$\to$EN & 57.09 (es) & 57.35 (da) & 56.95 & 57.29 & 57.23 & 57.31 \\
    \bottomrule
  \end{tabular}
  \caption{chrF score for the ensemble setting per model and ensemble size.}
  \label{tab:eval-chrf-ensemble}
\end{table*}

The results are summarized in \Cref{tab:ensemble-summary-chrf}. For each setting, we compare two matched models: (1) the best model after ensemble training, and (2) the best pre-ensemble model after the same amount of additional single-model training. This ensures that both models receive an equal number of extra training steps. To assess statistical significance, we perform paired t-tests on these matched performance pairs: mean baseline performance versus mean post-ensemble best model performance. Overall, the average improvement is $\mathbf{1.7}$ chrF from English (significant with $p=2 \cdot 10^{-4}, t=6.44$), $\mathbf{0.67}$ into English (significant with $p=0.00166, t=4.64$), and $\mathbf{1.19}$ on average (significant with $p=1 \cdot 10^{-5}, t=6.18$).
More detailed results are provided in \Cref{tab:eval-chrf-ensemble}. We also report BLEU and COMET scores in \Cref{tab:ensemble-summary-bleu,tab:ensemble-summary-comet}, and observe similar trends (average gains of $0.98$ BLUE, $1$ COMET\footnote{Normalized on a 0-100 scale.}).

\paragraph{Effect of Direction} As can be seen in \Cref{tab:ensemble-summary-chrf}, while ensemble almost always improves both translation directions, the magnitude of improvement is highly dependent on the direction of translation: for any model and ensemble size, the improvement was larger when translating into the non-English language - which likely received less data during pretraining. The average improvement difference is $1.03$ chrF (significant with $p=1.2 \cdot 10^{-4}$).

\begin{figure}[t]
  \centering
  \includegraphics[width=0.48\textwidth]{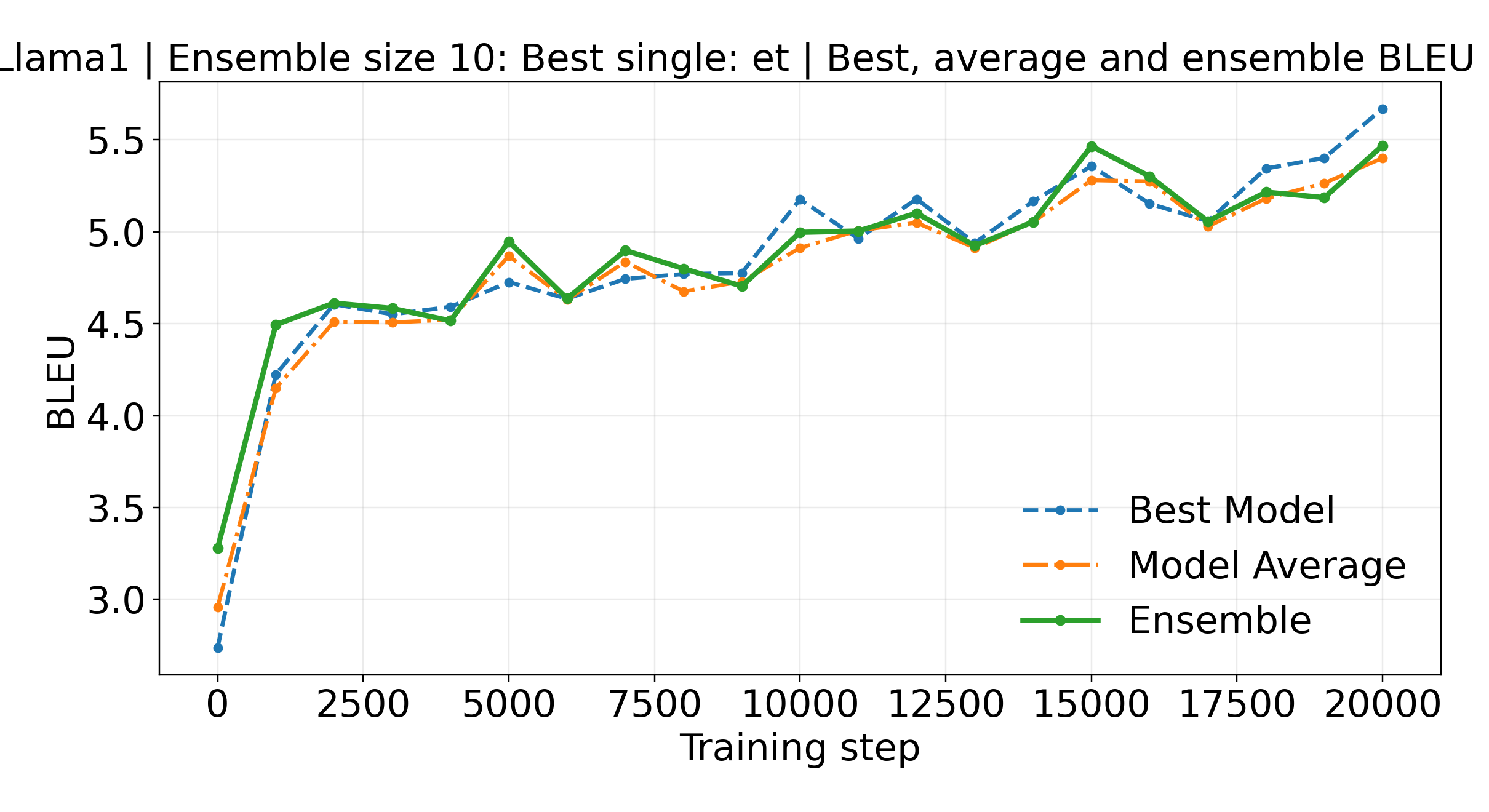}
  \caption{Llama-3.2-1B performance during ensemble training (10 models).}
  \label{fig:llama1-best-avg-ensemble}
\end{figure}

\begin{figure}[t]
  \centering
  \includegraphics[width=0.48\textwidth]{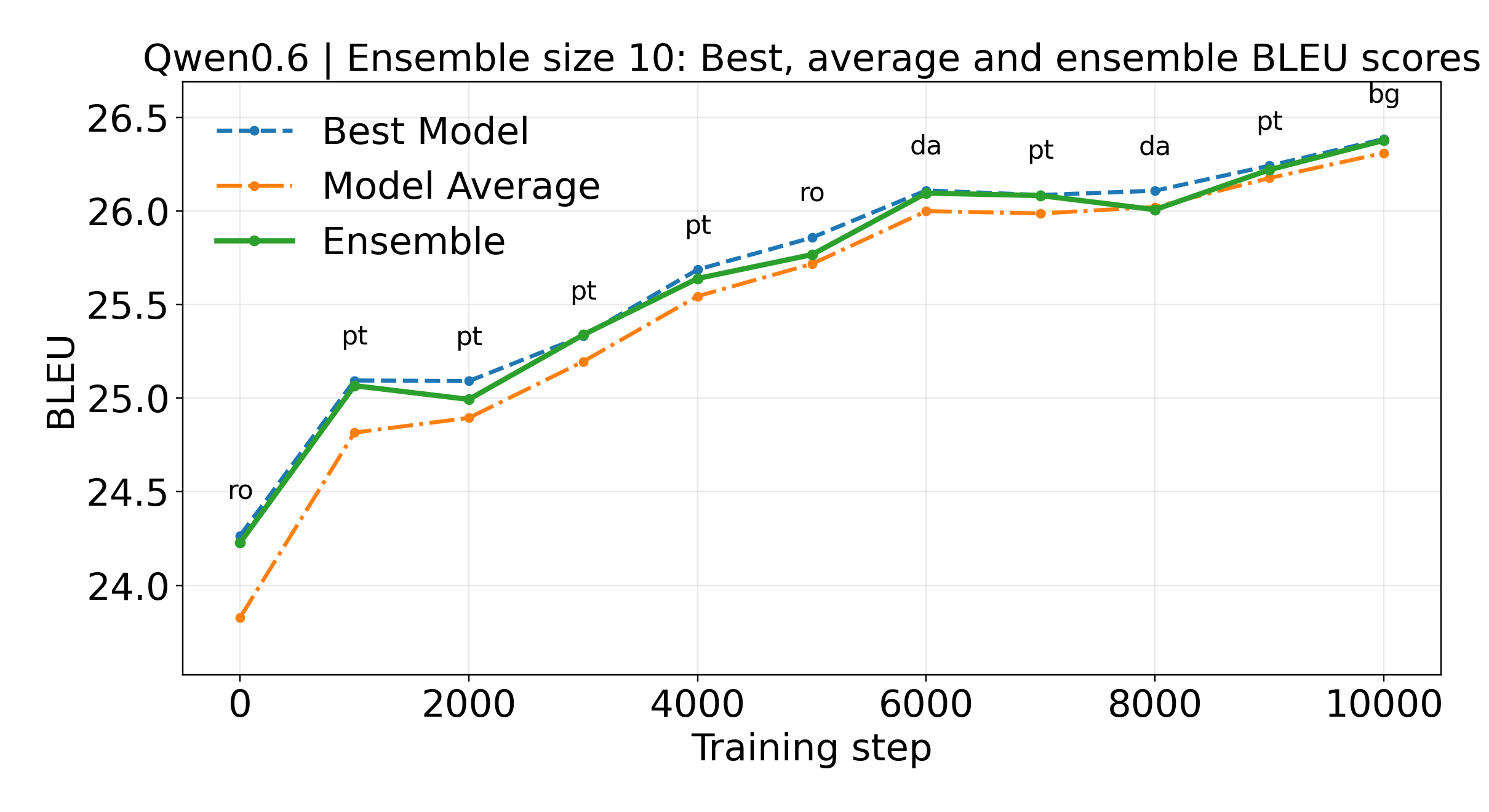}
  \caption{Qwen3-0.6B-Base performance during ensemble training (10 models).}
  \label{fig:qwen0.6-best-avg-ensemble}
\end{figure}

\begin{figure}[t]
  \centering
  \includegraphics[width=0.48\textwidth]{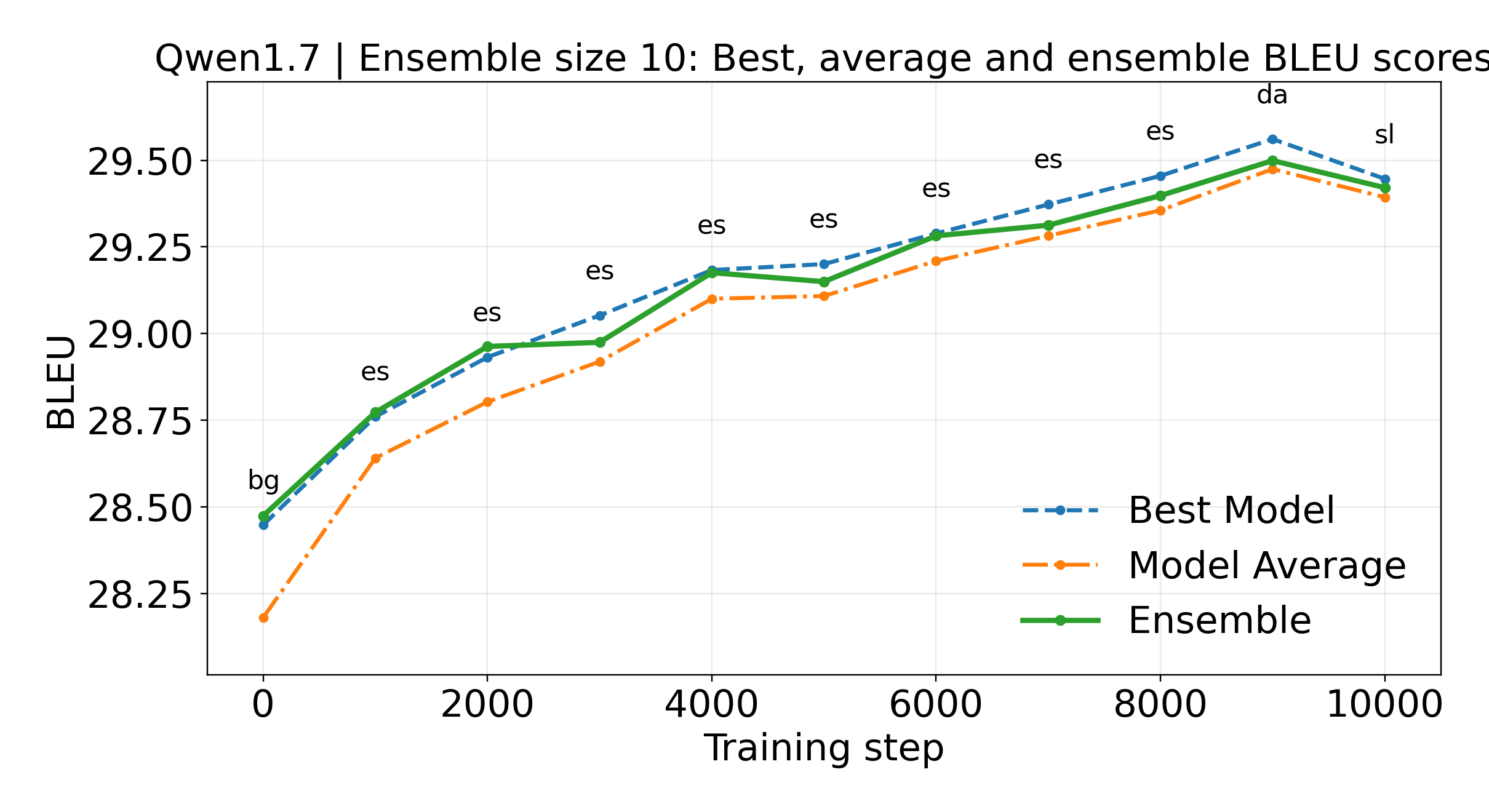}
  \caption{Qwen3-1.7B-Base performance during ensemble training (10 models).}
  \label{fig:qwen1.7-best-avg-ensemble}
\end{figure}


\paragraph{Ensemble Deployment} Apart from ensemble training, we analyze gains from ensembling models during inference. Looking at \Cref{tab:eval-chrf-ensemble}, results show that even without any training, ensemble provides a small but statistically significant improvement over the average of its models of $0.33$ chrF ($p=3.23 \cdot 10^{-4}, t=4.49$). However, the best individual model still outperforms ensemble deployment with an average difference of $0.346$ chrF (not significant). This raises a natural question: how can training on ensemble outputs improve the best model if the ensemble itself performs worse than that model?

\paragraph{Training Dynamics} The answer is that while ensemble performs worse than the best model, it still outperforms the average of the models. Training on ensemble-generated data improves most models, which in turn improves the ensemble and eventually benefits the best model. Indeed, as can be seen in \Cref{fig:llama1-best-avg-ensemble,fig:qwen0.6-best-avg-ensemble,fig:qwen1.7-best-avg-ensemble}, the best model is not fixed during training; this is expected as the worse models should improve more from training on the ensemble-generated data.

\section{Limitations}
While our experiments demonstrate statistically significant gains over non-ensemble methods with the same amount of training steps, more work is needed to test the limits and potential of ensemble-driven back-translation. 
This includes testing larger models, different ways to induce diversity among models (e.g. different base models), and evaluating architectures beyond generative LLMs, such as autoencoders \citep{artetxe2017unsupervised,chauhan2022fully}.
Of particular importance is combining our method with other methods that elicit translation capabilities, to test whether the gains are additive.
For example, denoising pretraining \citep{liu2020multilingual,garcia2020multilingual} or vocabulary extension techniques for the low-resource case \citep{chronopoulou2020reusing} could be tested. Particularly promising are strategies such as quality-aware pseudo-parallel selection \citep{xu2019improving,khatri2020filtering}, as the use of multiple models may make the filtering more efficient. We believe that future work should address these points.

\section{Conclusions}
We proposed an ensemble-driven framework in which multiple models collaboratively generate higher-quality pseudo-data through ensemble decoding. Experiments show consistent improvements over single-model training, with average gains of 1.19 chrF.

Future work could scale the approach to larger models, introduce alternative sources of model diversity, and combine the method with other unsupervised machine translation techniques, such as quality-aware filtering of synthetic data.

\clearpage
\bibliography{refs}

\clearpage

\appendix
\section{Appendix}

\subsection{Single-Model Tables}

\begin{table}[htbp]
\centering
\small
\begin{tabular}{|l|c|c|c|c|}
\hline
\makecell{Version \\ (3rd lang)} & PL->EN & EN->PL & 3rd->EN & EN->3rd \\
\hline
BG & 5.55 & 0.98 & 3.39 & 2.26 \\ \hline
CS & 4.76 & 1.73 & 2.62 & 1.82 \\ \hline
EL & 4.03 & 1.11 & 1.11 & 0.80 \\ \hline
ET & 4.17 & 1.30 & 2.57 & 1.52 \\ \hline
FI & 4.00 & 1.26 & 1.07 & 0.47 \\ \hline
HU & 4.53 & 1.17 & 3.43 & 2.50 \\ \hline
LT & 5.33 & 0.85 & 2.71 & 1.81 \\ \hline
LV & 5.19 & 1.55 & 4.19 & 2.91 \\ \hline
RO & 4.80 & 1.00 & 5.13 & 2.38 \\ \hline
SK & 4.34 & 1.36 & 2.00 & 0.90 \\ \hline
\end{tabular}
\caption{Final single-model Llama-3.2-1B 3000 sentence BLEU by direction (low-resource).}
\label{tab:llama1-all-single}
\end{table}

\begin{table}[htbp]
\centering
\small
\begin{tabular}{|l|c|c|c|c|}
\hline
\makecell{Version \\ (3rd lang)} & FR->EN & EN->FR & 3rd->EN & EN->3rd \\
\hline
BG & 26.17 & 22.13 & 29.99 & 20.90 \\ \hline
CS & 25.78 & 21.58 & 25.12 & 14.47 \\ \hline
DA & 25.67 & 21.39 & 21.88 & 15.61 \\ \hline
LV & 25.42 & 21.16 & 7.72 & 5.15 \\ \hline
NL & 25.45 & 21.62 & 20.13 & 14.76 \\ \hline
PL & 25.92 & 22.09 & 23.91 & 13.51 \\ \hline
PT & 26.13 & 22.36 & 24.65 & 21.01 \\ \hline
RO & 25.94 & 22.48 & 39.38 & 24.73 \\ \hline
SL & 25.59 & 21.57 & 18.97 & 11.94 \\ \hline
SV & 25.69 & 21.70 & 21.34 & 14.32 \\ \hline
\end{tabular}
\caption{Final single-model Qwen3-0.6B 3000 sentence BLEU by direction (high-resource).}
\label{tab:qwen0.6-all-single}
\end{table}

\begin{table}[htbp]
\centering
\small
\begin{tabular}{|l|c|c|c|c|}
\hline
\makecell{Version \\ (3rd lang)} & FR->EN & EN->FR & 3rd->EN & EN->3rd \\ \hline
BG & 30.29 & 26.62 & 41.31 & 31.47 \\ \hline
CS & 30.26 & 26.33 & 36.53 & 24.72 \\ \hline
DA & 29.74 & 26.06 & 29.45 & 21.95 \\ \hline
DE & 30.13 & 25.95 & 29.59 & 18.70 \\ \hline
ES & 30.08 & 26.27 & 32.32 & 30.57 \\ \hline
LV & 30.44 & 26.12 & 29.44 & 16.90 \\ \hline
NL & 30.00 & 26.23 & 25.37 & 20.22 \\ \hline
RO & 30.20 & 26.78 & 48.33 & 36.55 \\ \hline
SK & 30.18 & 26.12 & 38.67 & 25.92 \\ \hline
SL & 30.09 & 26.16 & 40.49 & 25.54 \\ \hline
SV & 30.01 & 26.28 & 29.02 & 21.78 \\ \hline
\end{tabular}
\caption{Final single-model Qwen3-1.7B 3000 sentence BLEU by direction (high-resource).}
\label{tab:qwen1.7-all-single}
\end{table}

\subsection{Training Details}
\label{sec:training-details}
Single-model training and evaluation was conducted on a single B200 GPU. Training took approximately 10 hours for Llama-3.2-1B (low-resource setting), 8 hours for Qwen-3-1.7B-Base (high-resource), and 5 hours for Qwen-3-0.6B-Base (high-resource).

Ensemble training and evaluation was performed on either one B200 GPU or two GPUs when additional memory was required. Training an ensemble of three models took approximately 24 hours for Llama-3.2-1B (low-resource) and Qwen-3-1.7B-Base (high-resource), and about 12 hours for Qwen-3-0.6B-Base (high-resource). Training time scales approximately linearly with the number of models in the ensemble.

\subsection{AI Assistant Use}
AI assistants were used solely for paraphrasing with ChatGPT and for identifying related work using Undermind. Independent manual searches for related work were also conducted. All AI-generated outputs were carefully reviewed and verified prior to being incorporated into the paper.

\subsection{Single Model Training Figures}

\begin{figure}[t]
  \centering
  \includegraphics[width=0.48\textwidth]{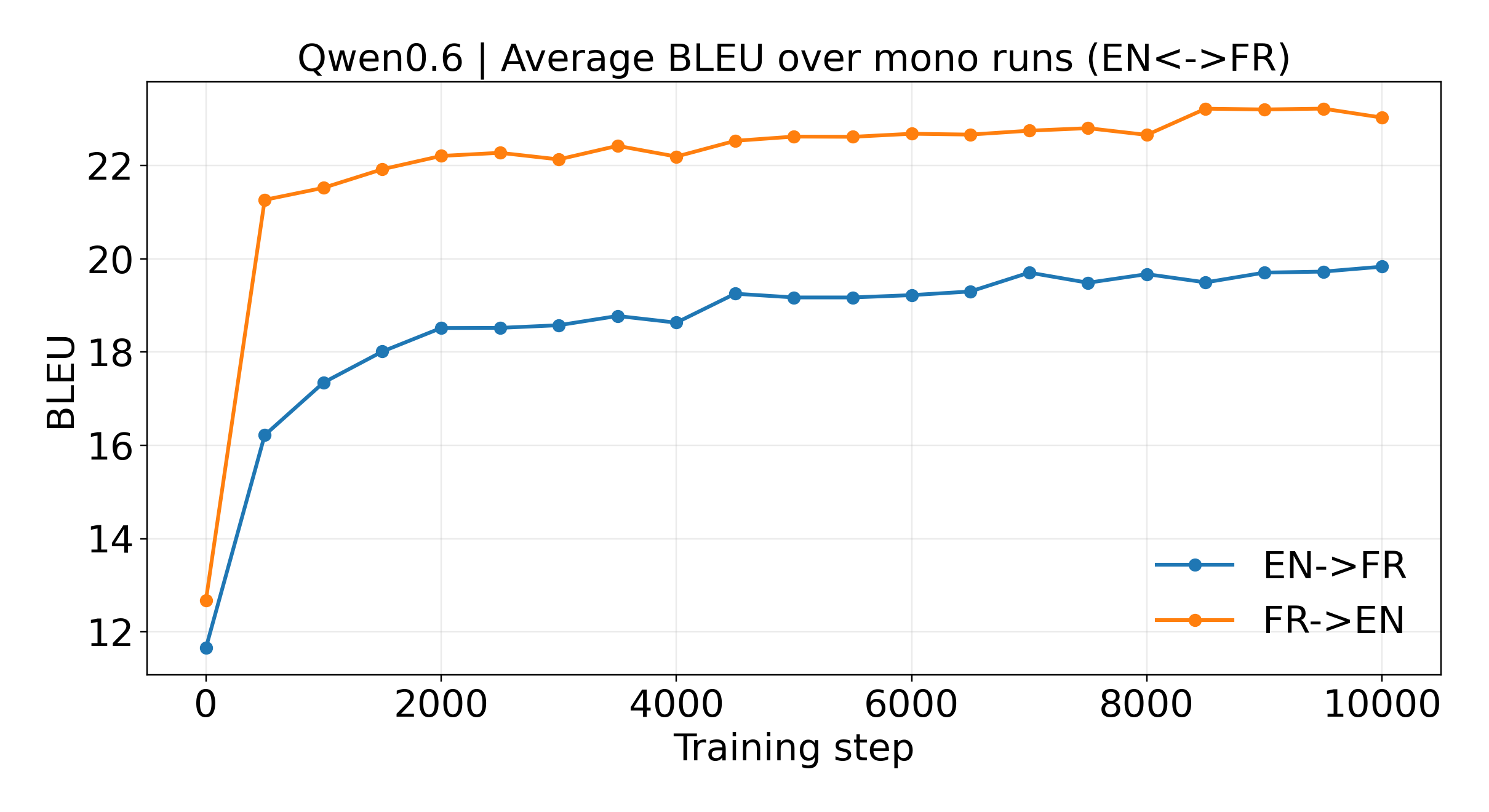}
  \caption{100 sentences BLEU for \texttt{Qwen3-0.6B-Base}: mean performance across all single models.}
  \label{fig:qwen0.6-single-avg}
\end{figure}

\begin{figure}[t]
  \centering
  \includegraphics[width=0.48\textwidth]{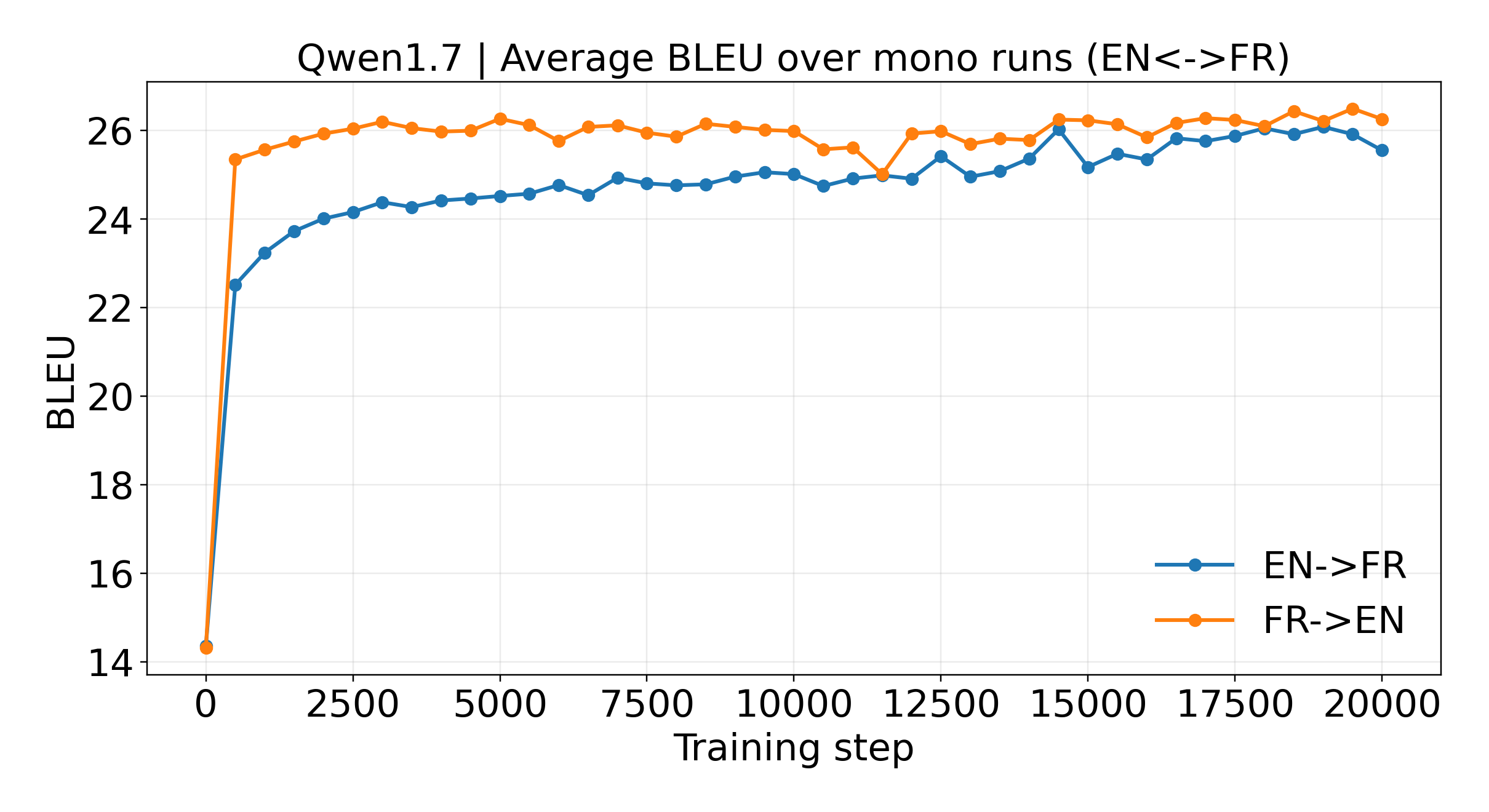}
  \caption{100 sentences BLEU for \texttt{Qwen3-1.7B-Base}: mean performance across all single models.}
  \label{fig:qwen1.7-single-avg}
\end{figure}

See \Cref{fig:qwen0.6-single-avg,fig:qwen1.7-single-avg}.

\subsection{Ensemble Score Tables}

See \Cref{tab:ensemble-summary-bleu,tab:ensemble-summary-comet,tab:eval-bleu-ensemble,tab:eval-comet-ensemble}.

\begin{table*}[t]
\centering

\begin{minipage}[t]{0.49\textwidth}
\centering
\scriptsize
\setlength{\tabcolsep}{3pt}
\captionof{table}{Best individual model BLEU score.}
\label{tab:ensemble-summary-bleu}
\resizebox{\linewidth}{!}{%
\begin{tabular}{lcccc}
\toprule
Model & \makecell{Ensemble \\ Size} & Direction & \makecell{Best \\ Model \\ Before \\ +Single\\Training} & \makecell{Best \\ Model \\ After \\Ensemble \\ Training} \\
\midrule

\multirow[c]{6}{*}[-1.2ex]{\makecell{Llama-1B \\ (low-\\resource)}}
& \multirow{2}{*}{3} & EN$\to$PL & 2.79 & 4.38 \\
&  & PL$\to$EN & 6.77 & 6.24 \\
\cmidrule(lr){2-5}
& \multirow{2}{*}{6} & EN$\to$PL & 2.79 & 4.23 \\
&  & PL$\to$EN & 6.77 & 6.33 \\
\cmidrule(lr){2-5}
& \multirow{2}{*}{10} & EN$\to$PL & 2.79 & 4.54 \\
&  & PL$\to$EN & 6.77 & 6.70 \\

\midrule

\multirow[c]{6}{*}[-1.2ex]{\makecell{Qwen-0.6B \\ (high-\\resource)}}
& \multirow{2}{*}{3} & EN$\to$FR & 22.82 & 25.31 \\
&  & FR$\to$EN & 26.40 & 27.48 \\
\cmidrule(lr){2-5}
& \multirow{2}{*}{6} & EN$\to$FR & 22.82 & 25.10 \\
&  & FR$\to$EN & 26.40 & 27.50 \\
\cmidrule(lr){2-5}
& \multirow{2}{*}{10} & EN$\to$FR & 23.76 & 25.30 \\
&  & FR$\to$EN & 26.40 & 27.49 \\

\midrule

\multirow[c]{6}{*}[-1.2ex]{\makecell{Qwen-1.7B \\ (high-\\resource)}}
& \multirow{2}{*}{3} & EN$\to$FR & 27.18 & 28.39 \\
&  & FR$\to$EN & 30.49 & 30.70 \\
\cmidrule(lr){2-5}
& \multirow{2}{*}{6} & EN$\to$FR & 27.18 & 28.31 \\
&  & FR$\to$EN & 30.40 & 30.77 \\
\cmidrule(lr){2-5}
& \multirow{2}{*}{10} & EN$\to$FR & 27.11 & 28.32 \\
&  & FR$\to$EN & 30.49 & 30.65 \\

\midrule

\multirow[c]{3}{*}[0ex]{\textbf{Mean}}
& \multirow{3}{*}{\textbf{*}} 
& \textbf{EN$\boldsymbol{\to}$*} & \textbf{17.69} & \textbf{19.32} \\
& & \textbf{*$\boldsymbol{\to}$EN} & \textbf{21.21} & \textbf{21.54} \\
& & \textbf{Mean} & \textbf{19.45} & \textbf{20.43} \\

\bottomrule
\end{tabular}%
}
\end{minipage}
\hfill
\begin{minipage}[t]{0.49\textwidth}
\centering
\scriptsize
\setlength{\tabcolsep}{3pt}
\captionof{table}{Best individual model COMET score.}
\label{tab:ensemble-summary-comet}
\resizebox{\linewidth}{!}{%
\begin{tabular}{lcccc}
\toprule
Model & \makecell{Ensemble \\ Size} & Direction & \makecell{Best \\ Model \\ Before \\ +Single\\Training} & \makecell{Best \\ Model \\ After \\Ensemble \\ Training} \\
\midrule

\multirow[c]{6}{*}[-1.2ex]{\makecell{Llama-1B \\ (low-\\resource)}}
& \multirow{2}{*}{3} & EN$\to$PL & 0.1997 & 0.2206 \\
&  & PL$\to$EN & 0.2069 & 0.1982 \\
\cmidrule(lr){2-5}
& \multirow{2}{*}{6} & EN$\to$PL & 0.1997 & 0.2151 \\
&  & PL$\to$EN & 0.2069 & 0.2078 \\
\cmidrule(lr){2-5}
& \multirow{2}{*}{10} & EN$\to$PL & 0.1997 & 0.2376 \\
&  & PL$\to$EN & 0.2069 & 0.2092 \\

\midrule

\multirow[c]{6}{*}[-1.2ex]{\makecell{Qwen-0.6B \\ (high-\\resource)}}
& \multirow{2}{*}{3} & EN$\to$FR & 0.6382 & 0.6766 \\
&  & FR$\to$EN & 0.7967 & 0.8163 \\
\cmidrule(lr){2-5}
& \multirow{2}{*}{6} & EN$\to$FR & 0.6382 & 0.6461 \\
&  & FR$\to$EN & 0.7967 & 0.7992 \\
\cmidrule(lr){2-5}
& \multirow{2}{*}{10} & EN$\to$FR & 0.6491 & 0.6789 \\
&  & FR$\to$EN & 0.8057 & 0.8180 \\

\midrule

\multirow[c]{6}{*}[-1.2ex]{\makecell{Qwen-1.7B \\ (high-\\resource)}}
& \multirow{2}{*}{3} & EN$\to$FR & 0.8048 & 0.8053 \\
&  & FR$\to$EN & 0.8862 & 0.8868 \\
\cmidrule(lr){2-5}
& \multirow{2}{*}{6} & EN$\to$FR & 0.8048 & 0.8063 \\
&  & FR$\to$EN & 0.8862 & 0.8860 \\
\cmidrule(lr){2-5}
& \multirow{2}{*}{10} & EN$\to$FR & 0.8048 & 0.8026 \\
&  & FR$\to$EN & 0.8862 & 0.8872 \\

\midrule

\multirow[c]{3}{*}[0ex]{\textbf{Mean}}
& \multirow{3}{*}{\textbf{*}} 
& \textbf{EN$\boldsymbol{\to}$*} & \textbf{0.5488} & \textbf{0.5655} \\
& & \textbf{*$\boldsymbol{\to}$EN} & \textbf{0.6309} & \textbf{0.6343} \\
& & \textbf{Mean} & \textbf{0.5899} & \textbf{0.5999} \\

\bottomrule
\end{tabular}%
}
\end{minipage}

\end{table*}

\begin{table*}[t]
  \centering
  \small
  \setlength{\tabcolsep}{4pt} 
  \begin{tabular}{cclcccccc}
    \toprule
    \makecell{Model Name} & \makecell{Ensemble \\ Size} & Direction & \makecell{Best Model \\ Before} &  \makecell{Best Model \\ After} &  \makecell{Avg. Models \\ Before} &  \makecell{Avg. Models \\ After} &  \makecell{Ensemble \\ Before} &  \makecell{Ensemble \\ After} \\

    \midrule
    
    \multirow{6}{*}[-1.2ex]{\texttt{Llama-3.2-1B}}
& \multirow{2}{*}[-0.3ex]{3}
    & EN$\to$PL & 1.70 (cs) & 4.38 (cs) & 1.24 & 4.15 & 1.87 & 4.03 \\
    & & PL$\to$EN & 5.57 (bg) & 6.24 (cs) & 4.79 & 5.85 & 5.00 & 5.61 \\

    \cmidrule(lr){2-9}

    & \multirow{2}{*}[-0.3ex]{6}
    & EN$\to$PL & 1.75 (cs) & 4.23 (et) & 1.24 & 4.19 & 1.55 & 4.17 \\
    & & PL$\to$EN & 5.51 (bg) & 6.33 (hu) & 4.50 & 6.23 & 5.07 & 6.26 \\

    \cmidrule(lr){2-9}

    & \multirow{2}{*}[-0.3ex]{10}
    & EN$\to$PL & 1.73 (cs) & 4.54 (et) & 1.22 & 4.36 & 1.69 & 4.36 \\
    & & PL$\to$EN & 5.61 (bg) & 6.70 (hu) & 4.69 & 6.46 & 4.87 & 6.53 \\
    \midrule

    \multirow{6}{*}[-1.2ex]{\texttt{Qwen3-0.6B}} & \multirow{2}{*}[-0.3ex]{3}
    & EN$\to$FR & 22.12 (bg) & 25.31 (bg) & 21.73 & 25.24 & 21.94 & 25.30 \\
    & & FR$\to$EN & 26.25 (bg) & 27.48 (bg) & 25.89 & 27.41 & 26.17 & 27.40 \\
    \cmidrule(lr){2-9}

    & \multirow{2}{*}[-0.3ex]{6}
    & EN$\to$FR & 22.28 (bg) & 25.10 (bg) & 21.95 & 23.97 & 22.38 & 25.09 \\
    & & FR$\to$EN & 26.02 (bg) & 27.50 (bg) & 25.72 & 26.83 & 26.08 & 27.33 \\
    \cmidrule(lr){2-9}

    & \multirow{2}{*}[-0.3ex]{10}
    & EN$\to$FR & 22.61 (ro) & 25.30 (pt) & 21.84 & 25.16 & 22.22 & 25.29 \\
    & & FR$\to$EN & 26.27 (bg) & 27.49 (ro) & 25.82 & 27.44 & 26.24 & 27.58 \\
    \midrule

    \multirow{6}{*}[-1.2ex]{\texttt{Qwen3-1.7B}} & \multirow{2}{*}[-0.3ex]{3}
    & EN$\to$FR & 26.61 (bg) & 28.39 (da) & 26.31 & 28.32 & 26.44 & 28.26 \\
    & & FR$\to$EN & 30.31 (bg) & 30.70 (bg) & 30.08 & 30.66 & 30.21 & 30.80 \\
    \cmidrule(lr){2-9}

    & \multirow{2}{*}[-0.3ex]{6}
    & EN$\to$FR & 26.52 (bg) & 28.31 (da) & 26.20 & 28.20 & 26.49 & 28.11 \\
    & & FR$\to$EN & 30.43 (lv) & 30.77 (bg) & 30.15 & 30.70 & 30.36 & 30.73 \\
    \cmidrule(lr){2-9}

    & \multirow{2}{*}[-0.3ex]{10}
    & EN$\to$FR & 26.60 (ro) & 28.32 (es) & 26.25 & 28.22 & 26.58 & 28.26 \\
    & & FR$\to$EN & 30.32 (bg) & 30.65 (nl) & 30.11 & 30.57 & 30.37 & 30.51 \\
    \midrule
  \end{tabular}
  \caption{BLEU score for the ensemble setting per model and ensemble size.}
  \label{tab:eval-bleu-ensemble}
\end{table*}

\begin{table*}[t]
  \centering
  \small
  \setlength{\tabcolsep}{4pt}
  \begin{tabular}{cclcccccc}
    \toprule
    \makecell{Model Name} & \makecell{Ensemble \\ Size} & Direction & \makecell{Best Model \\ Before} & \makecell{Best Model \\ After} & \makecell{Avg. Models \\ Before} & \makecell{Avg. Models \\ After} & \makecell{Ensemble \\ Before} & \makecell{Ensemble \\ After} \\
    \midrule
    \multirow{6}{*}[-0.6ex]{\texttt{Llama-3.2-1B}} & \multirow{2}{*}[-0.3ex]{3} & EN$\to$PL & 0.1806 (cs) & 0.2206 (cs) & 0.1707 & 0.2196 & 0.1831 & 0.2218 \\
     &  & PL$\to$EN & 0.1937 (bg) & 0.1982 (cs) & 0.1779 & 0.1939 & 0.1788 & 0.1943 \\
    \cmidrule(lr){2-9}
     & \multirow{2}{*}[-0.3ex]{6} & EN$\to$PL & 0.1812 (cs) & 0.2151 (el) & 0.1647 & 0.2097 & 0.1747 & 0.2098 \\
     &  & PL$\to$EN & 0.1924 (bg) & 0.2078 (cs) & 0.1697 & 0.2034 & 0.1819 & 0.2021 \\
    \cmidrule(lr){2-9}
     & \multirow{2}{*}[-0.3ex]{10} & EN$\to$PL & 0.1815 (cs) & 0.2376 (hu) & 0.1617 & 0.2280 & 0.1759 & 0.2257 \\
     &  & PL$\to$EN & 0.1926 (bg) & 0.2092 (et) & 0.1723 & 0.2005 & 0.1801 & 0.1988 \\
    \midrule
    \addlinespace[0.3ex]
    \multirow{6}{*}[-0.6ex]{\texttt{Qwen3-0.6B}} & \multirow{2}{*}[-0.3ex]{3} & EN$\to$FR & 0.6131 (bg) & 0.6766 (da) & 0.6023 & 0.6759 & 0.6119 & 0.6765 \\
     &  & FR$\to$EN & 0.8003 (bg) & 0.8163 (da) & 0.7853 & 0.8150 & 0.7947 & 0.8135 \\
    \cmidrule(lr){2-9}
     & \multirow{2}{*}[-0.3ex]{6} & EN$\to$FR & 0.6132 (bg) & 0.6461 (sv) & 0.5957 & 0.6436 & 0.6068 & 0.6464 \\
     &  & FR$\to$EN & 0.7985 (bg) & 0.7992 (de) & 0.7832 & 0.7976 & 0.7937 & 0.7989 \\
    \cmidrule(lr){2-9}
     & \multirow{2}{*}[-0.3ex]{10} & EN$\to$FR & 0.6206 (pt) & 0.6789 (pt) & 0.5981 & 0.6739 & 0.6116 & 0.6755 \\
     &  & FR$\to$EN & 0.7976 (bg) & 0.8180 (pt) & 0.7825 & 0.8145 & 0.7945 & 0.8159 \\
    \midrule
    \addlinespace[0.3ex]
    \multirow{6}{*}[-0.6ex]{\texttt{Qwen3-1.7B}} & \multirow{2}{*}[-0.3ex]{3} & EN$\to$FR & 0.7925 (bg) & 0.8053 (cs) & 0.7867 & 0.8042 & 0.7901 & 0.8042 \\
     &  & FR$\to$EN & 0.8874 (bg) & 0.8868 (da) & 0.8824 & 0.8862 & 0.8846 & 0.8856 \\
    \cmidrule(lr){2-9}
     & \multirow{2}{*}[-0.3ex]{6} & EN$\to$FR & 0.7916 (bg) & 0.8063 (da) & 0.7851 & 0.8041 & 0.7907 & 0.8034 \\
     &  & FR$\to$EN & 0.8866 (bg) & 0.8860 (bg) & 0.8812 & 0.8853 & 0.8869 & 0.8856 \\
    \cmidrule(lr){2-9}
     & \multirow{2}{*}[-0.3ex]{10} & EN$\to$FR & 0.7924 (bg) & 0.8026 (es) & 0.7849 & 0.8003 & 0.7919 & 0.8027 \\
     &  & FR$\to$EN & 0.8872 (bg) & 0.8872 (sk) & 0.8813 & 0.8856 & 0.8864 & 0.8864 \\
    \bottomrule
  \end{tabular}
  \caption{COMET score for the ensemble setting per model and ensemble size.}
  \label{tab:eval-comet-ensemble}
\end{table*}

\end{document}